\documentclass[runningheads]{llncs}
\usepackage{eccv}

\usepackage{eccvabbrv}

\usepackage{graphicx}
\usepackage{booktabs}
\usepackage{amssymb}% http://ctan.org/pkg/amssymb
\usepackage{pifont}% http://ctan.org/pkg/pifont
\newcommand{\cmark}{\textcolor{green}{\ding{51}}}%
\newcommand{\xmark}{\textcolor{red}{\ding{55}}}%
\usepackage{comment}
\usepackage[export]{adjustbox}
\usepackage{diagbox}
\usepackage{slashbox}
\usepackage{enumitem}
\usepackage{pifont}
\usepackage{multirow}
\usepackage[accsupp]{axessibility}  % Improves PDF readability for those with disabilities.

\usepackage{hyperref}

\usepackage{orcidlink}

\title{Dense Hand-Object(HO) GraspNet with Full Grasping Taxonomy and Dynamics} 
\titlerunning{HOGraspNet with Full Grasping Taxonomy and Dynamics}

\author{Woojin Cho\inst{1},\, Jihyun Lee\inst{1},\, Minjae Yi\inst{1},\, Minje Kim\inst{1},\, Taeyun Woo\inst{1},\, Donghwan Kim\inst{1},\, Taewook Ha\inst{1},\, Hyokeun Lee\inst{3},\, Je-Hwan Ryu\inst{4},\,\\ Woontack Woo\inst{1},\, Tae-Kyun Kim\inst{1,2}}
\authorrunning{Cho et al.}

\institute{KAIST\inst{1},\, Imperial College London\inst{2},\, Kwangwoon University\inst{3},\, Surromind\inst{4}}
\begin{document}
\maketitle
\vspace{-2\baselineskip}
\begin{abstract}
Existing datasets for 3D hand-object interaction are limited either in the data cardinality, data variations in interaction scenarios, or the quality of annotations. In this work, we present a comprehensive new training dataset for hand-object interaction called HOGraspNet. It is the only real dataset that captures full grasp taxonomies, providing grasp annotation and wide intraclass variations. Using grasp taxonomies as atomic actions, their space and time combinatorial can represent complex hand activities around objects. We select 22 rigid objects from the YCB dataset and 8 other compound objects using shape and size taxonomies, ensuring coverage of all hand grasp configurations. The dataset includes diverse hand shapes from 99 participants aged 10 to 74, continuous video frames, and a 1.5M RGB-Depth of sparse frames with annotations. It offers labels for 3D hand and object meshes, 3D keypoints, contact maps, and \emph{grasp labels}. Accurate hand and object 3D meshes are obtained by fitting the hand parametric model (MANO) and the hand implicit function (HALO) to multi-view RGBD frames, with the MoCap system only for objects. Note that HALO fitting does not require any parameter tuning, enabling scalability to the dataset's size with comparable accuracy to MANO. We evaluate HOGraspNet on relevant tasks: grasp classification and 3D hand pose estimation. The result shows performance variations based on grasp type and object class, indicating the potential importance of the interaction space captured by our dataset. The provided data aims at learning universal shape priors or foundation models for 3D hand-object interaction. Our dataset and code are available at https://hograspnet2024.github.io/.
\vspace{-0.5\baselineskip}
\end{abstract}

\begin{figure}[t]
  \centering
  \includegraphics[width=0.6\linewidth]{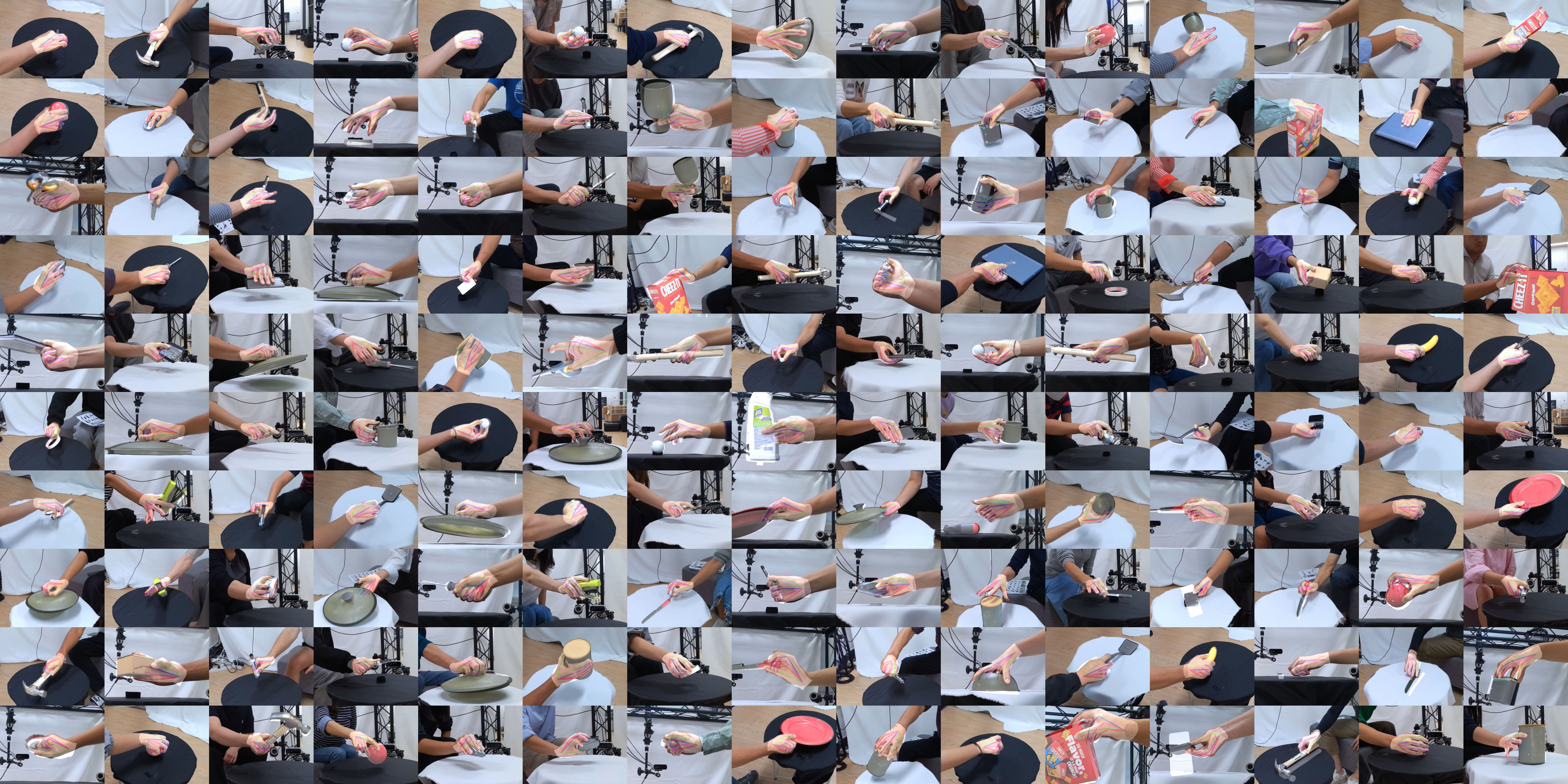}
\includegraphics[width=0.39\linewidth]{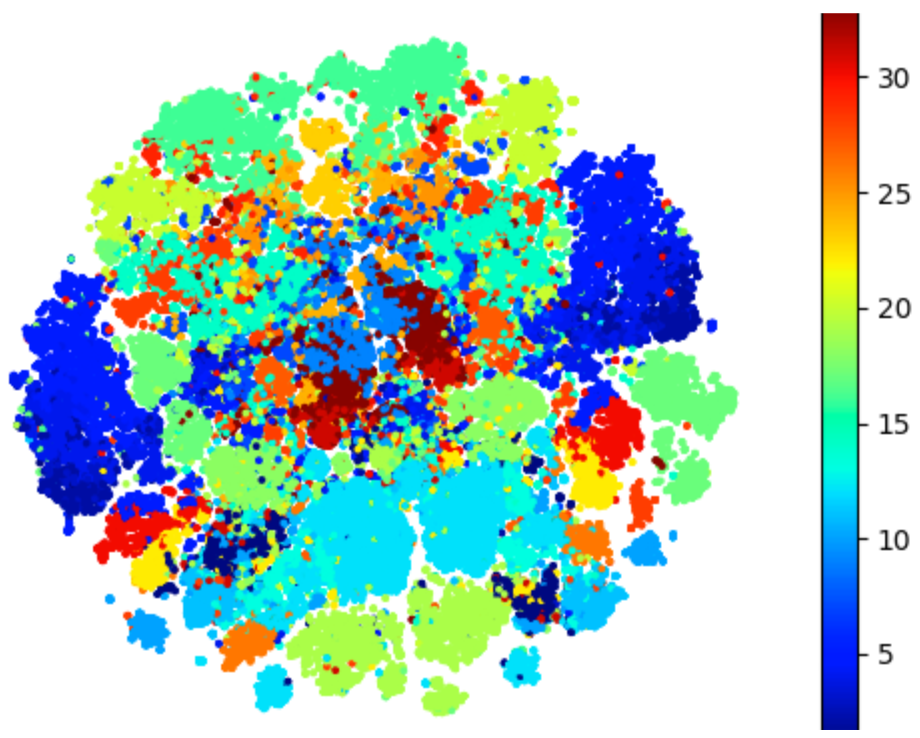}
   \caption{\textbf{(left) Diverse samples in HOGraspNet (best viewed with zoom-in).} HOGraspNet captures all hand-object grasp taxonomies with high-quality 3D annotations. \textbf{(right) Grasp Taxonomy t-SNE.} It covers well the grasp taxonomy space with intra-class variations.}
   \vspace{-2.0\baselineskip} 
\end{figure}
\label{fig:thumbnails}

\vspace{-2\baselineskip}

\section{Introduction}
\label{sec:intro}
\vspace{-0.5\baselineskip}

The importance of modeling and inferring 3D hand-object interactions is growing. While earlier works focused on single object instances~\cite{zimmermann2019freihand, zimmermann2021contrastive, simon2017hand, zhang2021hand, gomez2019large, yuan2017bighand2, tang2014latent, tompson2014real, qian2014realtime, xu2013efficient, yu2021local, zimmermann2017learning, zhang20163d, sridhar2013interactive}, recent efforts have been made on multiple 3D objects and their complex interactions~\cite{hampali2020honnotate, chao2021dexycb, hasson2019learning, mueller2017real, garcia2018first, brahmbhatt2020contactpose, taheri2020grab, fan2023arctic, corona2020ganhand, moon2020interhand2, Joo_2015_ICCV, singleshotmultiperson2018, singleshotmultiperson2018, joo2020eft, zheng2021deepmulticap, BEV, guo2022multi, yin2023hi4d, fan2023arctic
}. The human hand is the most dexterous and important testbed, and its research is extendable to human bodies or faces in similar articulated and deformable categories. We observe a few new benchmarks on hand-object interaction each latest year. However, existing datasets are limited either in the cardinality of data, the amount of data variations in hands/objects, or the quality of annotations. See Tab.~\ref{tab:comparisionHandObject}, where HO3D~\cite{hampali2020honnotate} and DexYCB~\cite{chao2021dexycb} include only 15 and 14 (out of 33) grasping taxonomies respectively. YCB Affordance~\cite{corona2020ganhand} is only an existing benchmark that represents all grasp taxonomies and provides grasp labels, but synthetic; ARCTIC places visible markers on hands in RGB images, and OakInk (the closest to ours) does not provide grasp labels with fewer subjects but more objects. More comparison with OakInk is shown in the t-SNE plot Fig.~\ref{fig:tsne}.

We introduce HOGraspNet, an extensive multi-view RGBD training dataset for hand, object, and their interaction with grasp annotations. Based on the existing hand grasping taxonomy \cite{feix2015grasp}, our design redefines 28 of the 33 grasps by merging geometrically similar or uncommon poses. Our dataset is the only real dataset covering all grasp taxonomies, including grasp labels and a wide range of intraclass variations. We exploit 22 rigid objects from the YCB dataset~\cite{calli2015ycb} with 8 other compound/articulated objects. As an example shown in Fig.~\ref{fig:allobjects}, 3 distinct hand grasps are performed for each object, totaling 90 interaction scenarios. Note that 30 objects are chosen enough to cover all grasp taxonomies, while textures and shapes beyond grasp areas can be synthetically augmented with 3D models. The dataset comprises a diverse range of hand identities from 99 participants aged 10 to 74. Overall, the dataset contains 1.5M RGB-Depth frames from 4 viewpoints, with annotations for 3D meshes, 3D keypoints, contact maps, and \emph{grasp labels}. We adopt the hand parametric model (MANO~\cite{romero2022embodied}) and the hand implicit function (HALO~\cite{karunratanakul2021skeleton}) individually to annotate the hand mesh. Presenting the novel annotation pipeline using the hand implicit function that requires simpler settings (i.e., less hyper-parameters) than MANO with accuracy and continuous shape representation. Considering the small objects in HOGraspNet, we utilize optical markers for MoCap only to obtain object 6D pose. We report experimental results of grasp classification, and SOTA hand-object 3D pose estimation methods. The new dataset demonstrates its comprehensiveness and potential. 

Further possibilities from the presented dataset are to 1) synthetically augment the data by changing backgrounds or object textures and shapes beyond grasp areas, 2) provide environments for learning a grasping agent in physical simulators, 3) extend to non-grasping actions, e.g., pushing, throwing, or deformed processes of non-rigid objects by hand. We hope the new dataset serves as a basis for understanding and modeling diverse inference models of hand-object interactions and that learned knowledge applies to human-object or human-human interactions.

\begin{table}[t]
\caption{\textbf{Comparison of hand-object interaction datasets.} Interaction info is the key criteria utilized in terms of hand-object interaction.}
\resizebox{\textwidth}{!}{
\begin{tabular}{c|cccccc|cccccccc}
\hline
Dataset &
  Type &
  \#image &
  \#views &
  \#obj &
  \#subj &
  \begin{tabular}[c]{@{}c@{}}\#Grasps \\in \cite{feix2015grasp}\end{tabular} &
  \begin{tabular}[c]{@{}c@{}}Real\end{tabular} &
  \begin{tabular}[c]{@{}c@{}}Video\end{tabular} &
  \begin{tabular}[c]{@{}c@{}}Marker-less\\ hand\end{tabular} &
  \begin{tabular}[c]{@{}c@{}}Dynamic\\ interaction\end{tabular} &
  \begin{tabular}[c]{@{}c@{}}Hand-obj\\ contactmap\end{tabular} &
  \begin{tabular}[c]{@{}c@{}}Grasp\\ variation\end{tabular} &
  \begin{tabular}[c]{@{}c@{}}Grasp\\ annotation\end{tabular} &
  \begin{tabular}[c]{@{}c@{}}Interaction\\ info.\end{tabular} \\ \hline
Obman(CVPR19)\cite{hasson19_obman} & RGBD  & 154k & 1 & 3k        & 20 &      & \xmark & \xmark& \cmark & \xmark & \xmark & \xmark & \xmark &       \\
YCB-Affordance(CVPR20)\cite{corona2020ganhand} & RGB  & 133k & 1 & 58        & - &  100\%   & \xmark & \cmark & \cmark & \xmark & \xmark & \cmark & \cmark &  Grasp     \\
FreiHAND(ICCV18)\cite{zimmermann2019freihand} & RGB   & 37k  & 8 & 2         & 32 &      & \cmark & \xmark & \cmark & \xmark & \xmark & \xmark & \xmark &       \\
MOW(ICCV21)\cite{cao2021reconstructing} & RGB   & 500  & 1 & $\sim$500 & -  & 82\% & \cmark & \xmark & \cmark & \xmark & \xmark & \xmark & \xmark &       \\
DexYCB(CVPR21)\cite{chao2021dexycb} & RGBD  & 582k & 8 & 20        & 10 & 42\% & \cmark & \cmark & \cmark & \xmark & \xmark & \xmark & \xmark &       \\
FPHA(CVPR18)\cite{garcia2018first} & RGBD  & 105k & 1 & 4         & 6  &      & \cmark & \cmark & \xmark & \cmark & \xmark & \xmark & \xmark &  Action     \\
HO3D(CVPR20)\cite{hampali2020honnotate} & RGBD  & 78k  & 1 & 10        & 10 & 45\% & \cmark & \cmark & \cmark & \cmark & \xmark & \xmark & \xmark &       \\
SHOWMe(ICCVW23)\cite{swamy2023showme} & RGBD  & 87k  & 1 & 42        & 15 & 61\% & \cmark & \cmark & \cmark & \cmark & \xmark & \xmark & \xmark &       \\ 
ContactPose(ECCV20)\cite{brahmbhatt2020contactpose} & RGBD  & 2.9M & 3 & 25        & 50 &      & \cmark & \cmark& \cmark & \cmark & \cmark & \xmark & \xmark & Intent      \\
H2O(ICCV21)\cite{kwon2021h2o} & RGBD  & 571k & 5 & 8         & 4  &      & \cmark & \cmark & \cmark & \cmark & \cmark & \xmark & \xmark &       \\
ARCTIC(CVPR23)\cite{fan2023arctic}      & RGBD  & 2.1M & 9 & 10        & 9  &      & \cmark & \cmark & \xmark & \cmark & \cmark & \cmark & \xmark & Intent \\ 
OakInk(CVPR22)\cite{yang2022oakink}      & RGBD & 230k & 4 & 100       & 12 &      & \cmark & \cmark & \cmark & \cmark & \cmark & \cmark & \xmark &       \\ \hline
Ours              & RGBD   & 1.5M & 4 & 30        & 99 & 85\% & \cmark & \cmark & \cmark & \cmark & \cmark & \cmark & \cmark &  Grasp    \\ \hline
\end{tabular}
}
\label{tab:comparisionHandObject}
\vspace{-1\baselineskip}
\end{table}

\vspace{-0.5\baselineskip}
\section{Survey on Interaction Datasets}
\label{sec:literature_survey}

This section provides a comprehensive overview of existing datasets on the interaction of 3D shapes, i.e., single hand, hand-object, hand-hand, human-object, and human-human. We also briefly discuss the existing literature on hand-object reconstruction, which is used for benchmarking our dataset (in Section \ref{sec:experiments}). Further survey results are available in the supplementary materials.

\vspace{-\baselineskip}

\subsubsection{Single-Hand Datasets.}
Earlier research efforts to build a hand dataset have focused on capturing single hands from RGB~\cite{zimmermann2019freihand, zimmermann2021contrastive, simon2017hand, zhang2021hand, gomez2019large}, depth~\cite{yuan2017bighand2, tang2014latent, tompson2014real, qian2014realtime, xu2013efficient}, or RGBD~\cite{yu2021local, zimmermann2017learning, zhang20163d, sridhar2013interactive}, stimulating various learning-based methods for hand reconstruction~\cite{pavlakos2024reconstructing, caramalau2021active, park20203d, garcia2020physics, lin2021mesh}. These datasets can be categorized via three characteristics: (1) whether the captured hand frames are synthetic~\cite{simon2017hand, zimmermann2017learning} or real~\cite{zimmermann2019freihand, zimmermann2021contrastive, simon2017hand, gomez2019large, zhang2021hand, qian2014realtime, tang2014latent, tompson2014real, xu2013efficient, yuan2017bighand2, sridhar2013interactive, yu2021local, zhang20163d}, (2) whether the hand is annotated as sparse keypoints~\cite{zimmermann2019freihand, simon2017hand, gomez2019large, zhang2021hand, zhang2021hand, tang2014latent, tompson2014real, xu2013efficient, yuan2017bighand2, zimmermann2017learning} or mesh~\cite{zimmermann2021contrastive, yu2021local}, and (3) whether the annotation is obtained via marker-based~\cite{gomez2019large, qian2014realtime, tang2014latent, yuan2017bighand2} or marker-less system~\cite{simon2017hand, zhang2021hand, zimmermann2021contrastive, zimmermann2019freihand, tompson2014real, xu2013efficient, sridhar2013interactive, yu2021local, zhang20163d, zimmermann2017learning}. More recently, various hand datasets aim to capture hands in interaction with an object~\cite{hampali2020honnotate, chao2021dexycb, hasson2019learning, mueller2017real, garcia2018first, brahmbhatt2020contactpose, taheri2020grab, fan2023arctic, swamy2023showme, corona2020ganhand, damen2022rescaling, brahmbhatt2019contactdb} or another hand~\cite{moon2024dataset, moon2020interhand2, lin2023handdiffuse, zuo2023reconstructing, tzionas2016capturing}. Since the goal of our work is to collect a dataset that comprehensively captures hand-object interactions, we focus on discussing the existing hand-object datasets in the following.

\vspace{-1\baselineskip}
\subsubsection{Hand-Object Datasets.} 
Recently, various hand-object datasets~\cite{hampali2020honnotate, chao2021dexycb, hasson2019learning, mueller2017real, garcia2018first, brahmbhatt2020contactpose, taheri2020grab, fan2023arctic, swamy2023showme, corona2020ganhand, damen2022rescaling, brahmbhatt2019contactdb} have been proposed. Regarding \textbf{(1) annotation method}, most of the earlier datasets collect synthetic RGB and/or depth images rendered from a parametric hand model (MANO~\cite{romero2022embodied}) and template object models~\cite{hasson2019learning, mueller2017real, corona2020ganhand}, or collect real images with markers~\cite{brahmbhatt2020contactpose, taheri2020grab, fan2023arctic} or magnetic sensors~\cite{garcia2018first} to obtain hand annotations. However, these samples lack realism due to the rendering of synthetic models or the presence of visible sensors. Thus, many recent datasets use a markerless system to fit the MANO model to RGB-D images captured in a multi-view setup while using a minimal number of markers to obtain object poses~\cite{hampali2020honnotate, chao2021dexycb, swamy2023showme}. Our work also follows such marker-less capture system to provide MANO-based hand annotations while additionally fitting an implicit function-based hand model (HALO~\cite{karunratanakul2021skeleton}) to provide supplementary hand shape information. Regarding the \textbf{(2) characteristics of captured data}, existing datasets are limited either in data cardinality, the number of object categories or hand identities, or interaction taxonomies 
(please refer to Tab.~\ref{tab:comparisionHandObject}). For example, HO3D~\cite{hampali2020honnotate} and DexYCB~\cite{chao2021dexycb} (which are the most widely used hand-object datasets) only consider 10 object categories and capture 10 and 20 hand identities, respectively. While ObMan~\cite{hasson2019learning} and SHOWMe~\cite{swamy2023showme} capture more diverse object categories, they are limited in the number of hand identities (20 and 15, respectively) and the data cardinality (154K and 87K, respectively). ARCTIC\cite{fan2023arctic} and OakInk\cite{yang2022oakink} are recently proposed datasets that capture dexterous interactions between hands and objects, containing a range of motion variations. However, they do not cover the diverse grasp poses for each object, as they instruct participants to assume poses based on their intent to interact with the object. Our work aims to collect a training dataset that is more comprehensive in terms of interaction scenarios based on grasps, object categories, hand identities, and data cardinality. We also note that most of the existing datasets do not provide a grasping type of each sample, which can further provide a useful prior for the captured hand-object interaction~\cite{corona2020ganhand, goyal2022human, liu2014taxonomy}. Our work also carefully identifies a taxonomy of 33 grasping types and provides grasping type annotation for each sample. For comparison with other datasets, we conducted a thorough survey to ascertain the number of grasp classes present among the 33 grasp taxonomies in Feix et al.\cite{feix2015grasp} and reported in Tab.~\ref{tab:comparisionHandObject}. The detailed results are provided in the supplementary material.

\vspace{-\baselineskip}
\subsubsection{Two-Hand Datasets.}
Similar to hand-object datasets, various interacting two-hand datasets have been proposed. HIC~\cite{tzionas2016capturing} and RGB2Hands~\cite{wang2020rgb2hands} are some of the earliest two-hand interaction datasets, but their data cardinality and interaction diversity are small compared to more recent datasets. InterHand2.6M~\cite{moon2020interhand2} is the most widely-used large-scale dataset, which captures interacting hands in a multi-view markerless motion capture setup. More recently, Re:InterHand~\cite{moon2024dataset} is proposed to capture more diverse two-hand interactions in terms of image appearances and interaction poses via the use of environment maps and hand relighting. TwoHand500K~\cite{zuo2023reconstructing} is another recently proposed dataset consisting of (1) real data captured using a marker-based system and (2) synthetic data obtained via combining poses randomly sampled from single-hand datasets. These datasets have inspired various methods for interacting two-hand reconstruction~\cite{li2022interacting, zhang2021interacting, lee2023im2hands} and generation~\cite{lee2024interhandgen, lin2023handdiffuse}.

\begin{table}[t]
\caption{\textbf{Comparison of Human-object interaction datasets.}}
\resizebox{\textwidth}{!}{
\centering
\begin{tabular}{c|cccccc|cccccccc}
\hline
Dataset &
  Type &
  \#image &
  \#views &
  \#obj &
  \#subj &
  \#Kinects &
  Label &
  \begin{tabular}[c]{@{}c@{}}Contact \\ annotation\end{tabular} &
  \begin{tabular}[c]{@{}c@{}}Whole body\\ interact.\end{tabular} &
  \begin{tabular}[c]{@{}c@{}}Marker-less\\ hand\end{tabular} &
  \begin{tabular}[c]{@{}c@{}}Natural\\ scene\end{tabular} &
  \begin{tabular}[c]{@{}c@{}}Scene\\interact.\end{tabular} &
  \begin{tabular}[c]{@{}c@{}}Dynamic\\hand\end{tabular} &
  \begin{tabular}[c]{@{}c@{}}Articulated\\ object\end{tabular} \\ \hline
EgoBody(ECCV22)\cite{zhang2022egobody}  & RGBD & 220K & 3$\sim$5 & 15 & 36 & 3$\sim$5 & SMPL-X & \xmark & \cmark & \cmark & \cmark & \cmark & \xmark & \xmark \\
GRAB(ECCV20)\cite{taheri2020grab} & Mesh    & 1.6M & -   & 51 & 10 & -   & SMPL-X & \cmark & \xmark & \xmark & \xmark & \xmark & \cmark & \xmark \\
ARCTIC(CVPR23)\cite{fan2023arctic}   & RGB  & 2.1M & 9 & 10 & 9  & -   & SMPL-X & \cmark & \xmark & \xmark & \xmark & \xmark & \cmark & \cmark \\
CHAIRS(ICCV23)\cite{jiang2023chairs}   & RGBD & 1.7M & 4   & 81 & 46 & -   & SMPL-X & \cmark & \cmark & \xmark & \xmark & \xmark & \xmark & \cmark \\
BEHAVE(CVPR22)\cite{bhatnagar22behave}   & RGBD & 15K  & 5   & 20 & 8  & -   & SMPL   & \cmark & \cmark & \xmark & \cmark & \xmark & \xmark & \xmark \\
InterCap(IJCV24)\cite{huang2024intercap} & RGBD & 67K  & 6   & 10 & 10 & 6   & SMPL-X & \cmark & \cmark & \cmark & \cmark & \xmark & \xmark & \xmark \\
PROX(ICCV19)\cite{PROX:2019}     & RGBD & 100K & 3   & 12 & 20 & 1   & SMPL-X & \cmark & \cmark & \cmark & \cmark & \cmark & \xmark & \xmark \\
RICH(CVPR22)\cite{huang2022rich}     & RGBD & 540K & 6$\sim$8 & 5  & 22 & 1   & SMPL-X & \cmark & \cmark & \cmark & \cmark & \cmark & \xmark & \xmark \\ \hline
\end{tabular}
}
\label{tab:comparisionHumanObject}
\vspace{-1\baselineskip}
\end{table}

\vspace{-0.8\baselineskip}
\subsubsection{Interacting Human Datasets.} 
In addition, research attention to interacting hands and several datasets contain interacting humans. These can be categorized as human-object interaction, human-human interaction, and human-scene interaction. \textbf{(1) Human-object interaction}: As demonstrated in Tab.~\ref{tab:comparisionHumanObject}, several datasets have been released to depict various aspects of human-object interactions. GRAB\cite{taheri2020grab} and ARCTIC\cite{fan2023arctic} propose datasets with 3d human mesh which focus on hand-object interaction. These three datasets \cite{bhatnagar22behave, jiang2023chairs, huang2024intercap} extend the interacting region to the whole body. Leveraging SMPL-X as mesh templates, \cite{bhatnagar22behave, fan2023arctic, huang2024intercap} contain contact supervision, and \cite{jiang2023chairs, fan2023arctic} focus on articulated objects. \textbf{(2) Human-human interaction}: PanopticStudio\cite{Joo_2015_ICCV}, MuCo-3DHP\cite{singleshotmultiperson2018} and MuPoTS-3D\cite{singleshotmultiperson2018} propose datasets with 3d human sparse keypoints. More recently, there has been increased interest on reconstructing 3d human mesh \cite{pumarola20193dpeople, leroy2020smply, Fieraru_2020_CVPR, Patel:CVPR:2021, joo2020eft, zheng2021deepmulticap, BEV, guo2022multi, yin2023hi4d}. The majority of them employ parametric models like SMPL\cite{SMPL:2015} or SMPL-X\cite{SMPL-X:2019}, except some datasets \cite{guo2022multi, pumarola20193dpeople, Patel:CVPR:2021, zheng2021deepmulticap, guo2022multi, yin2023hi4d} that provides textured scans, which is beneficial to represent geometric details. \textbf{(3) Human-scene interaction}: Certain datasets\cite{PROX:2019, huang2022rich, zhang2022egobody} broaden the scope of the interaction to include scenes, utilizing SMPL-X as a mesh template. PROX\cite{PROX:2019} and RICH\cite{huang2022rich} provide contact supervision between humans and scenes. RICH\cite{huang2022rich}, in particular, extends its scope to outdoor scenes, and EgoBody\cite{zhang2022egobody} provides motion text labels.

\vspace{-\baselineskip}
\subsubsection{Hand-Object Reconstruction.}
Reconstructing hand and object in interaction has been actively explored. Most of the recent works can be categorized into optimization or learning-based methods. Optimization-based methods~\cite{cao2021reconstructing, hasson2021towards,tse2022s,yang2021cpf,grady2021contactopt} typically fit MANO~\cite{romero2022embodied} hand and template object models based on contact or other physical constraints (e.g., attraction and repulsion~\cite{yang2021cpf}, friction~\cite{hu2022physical}). Learning-based methods~\cite{chen2021joint,liu2021semi,doosti2020hope,hasson2020leveraging,hasson2019learning,hampali2022keypoint,tekin2019h,chen2022alignsdf,chen2023gsdf,cho2018tracking,cho2020bare} directly regress hand and object poses via a neural network, while focusing on exploring an effective architecture for feature learning~\cite{chen2021joint,liu2021semi,doosti2020hope,hasson2020leveraging,hasson2019learning,hampali2022keypoint,tekin2019h} and/or shape representation~\cite{chen2023gsdf,chen2022alignsdf}. In our work, we benchmark the RGB-based reconstruction task on our HOGraspNet dataset using HFL-Net~\cite{hfl2022}, which is the most recent state-of-the-art method. 

\section{HOGraspNet}
\vspace{-0.5\baselineskip}

\subsection{Dataset Overview}
\vspace{-0.2\baselineskip}

\begin{figure}[t]
  \centering
  \includegraphics[width=0.8\linewidth]{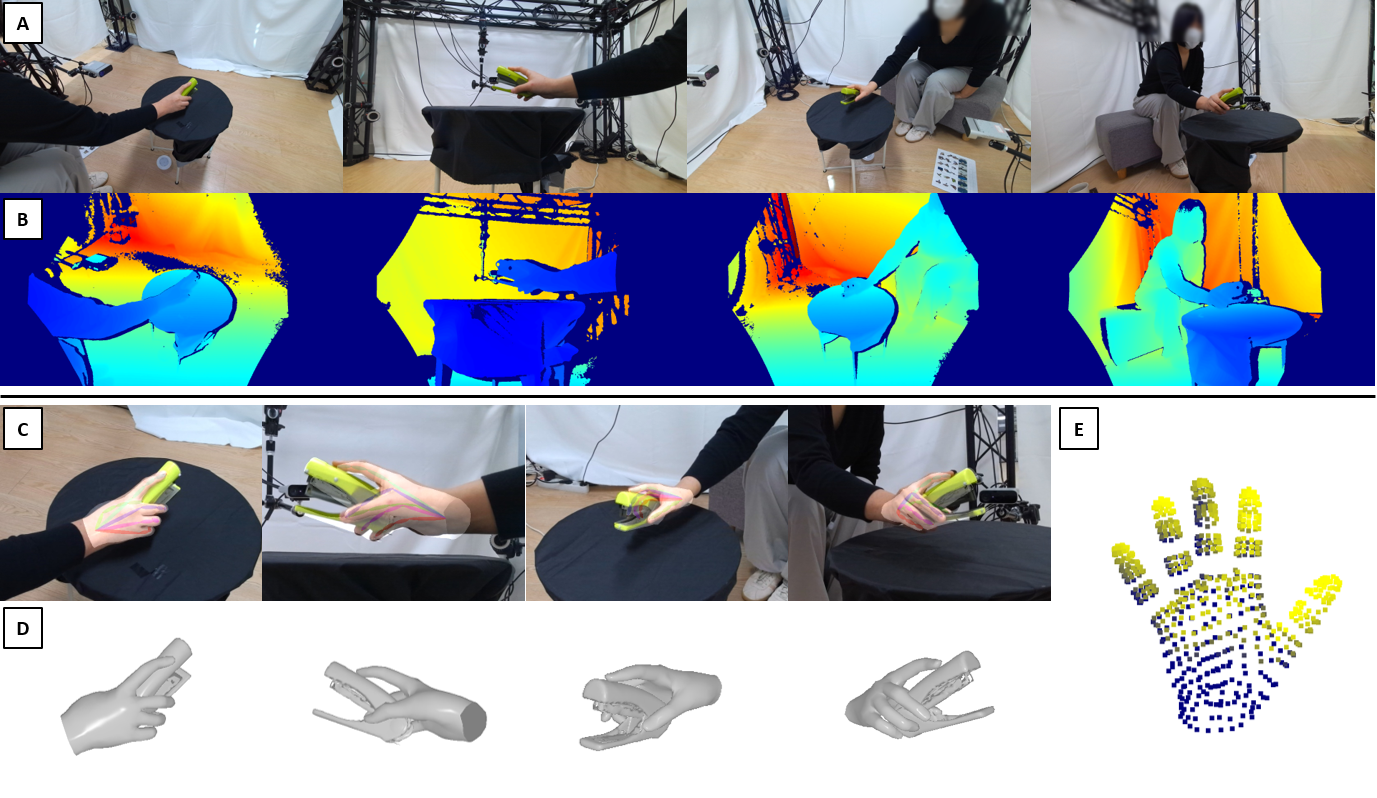}
   \caption{\textbf{Structure of HOGraspNet.} It captures diverse hand-object grasping at 4 different viewpoints. Example RGB images (A) and depth images (B) are shown, while the fitted hand and object meshes are visualized in (C) and (D). (E) shows the contact map.} 
\label{fig:datasOrganization}

\vspace{-\baselineskip}
\end{figure}

The dataset includes continuous video images and 1,489,112 annotated RGB-D frames, covering 28 hand grasp classes. We redefined the grasp classes by merging visually similar configurations (see supplementary) using 30 objects. The frames were captured at 4 distinct viewpoints and performed by 99 participants aged 10 to 74. Along with diverse hand shapes, a good scope of intra-class variations within each grasp class has been collected, which is important as a training dataset. Each RGB-D frame is annotated with the 3D hand pose for 21 joints and mesh, corresponding grasp class, 6D object pose, and contact map between the hand and object. Hand mesh models are obtained fitting MANO~\cite{romero2022embodied} and/or HALO~\cite{karunratanakul2021skeleton}, while all object mesh models (3d shapes and textures) are pre-scanned and provided. In Fig.~\ref{fig:datasOrganization}, we present examples of the data types included in the dataset.

\subsection{Object Categories and Grasp Taxonomy}

\begin{figure}[t]
  \centering
  \includegraphics[height=0.28\linewidth]{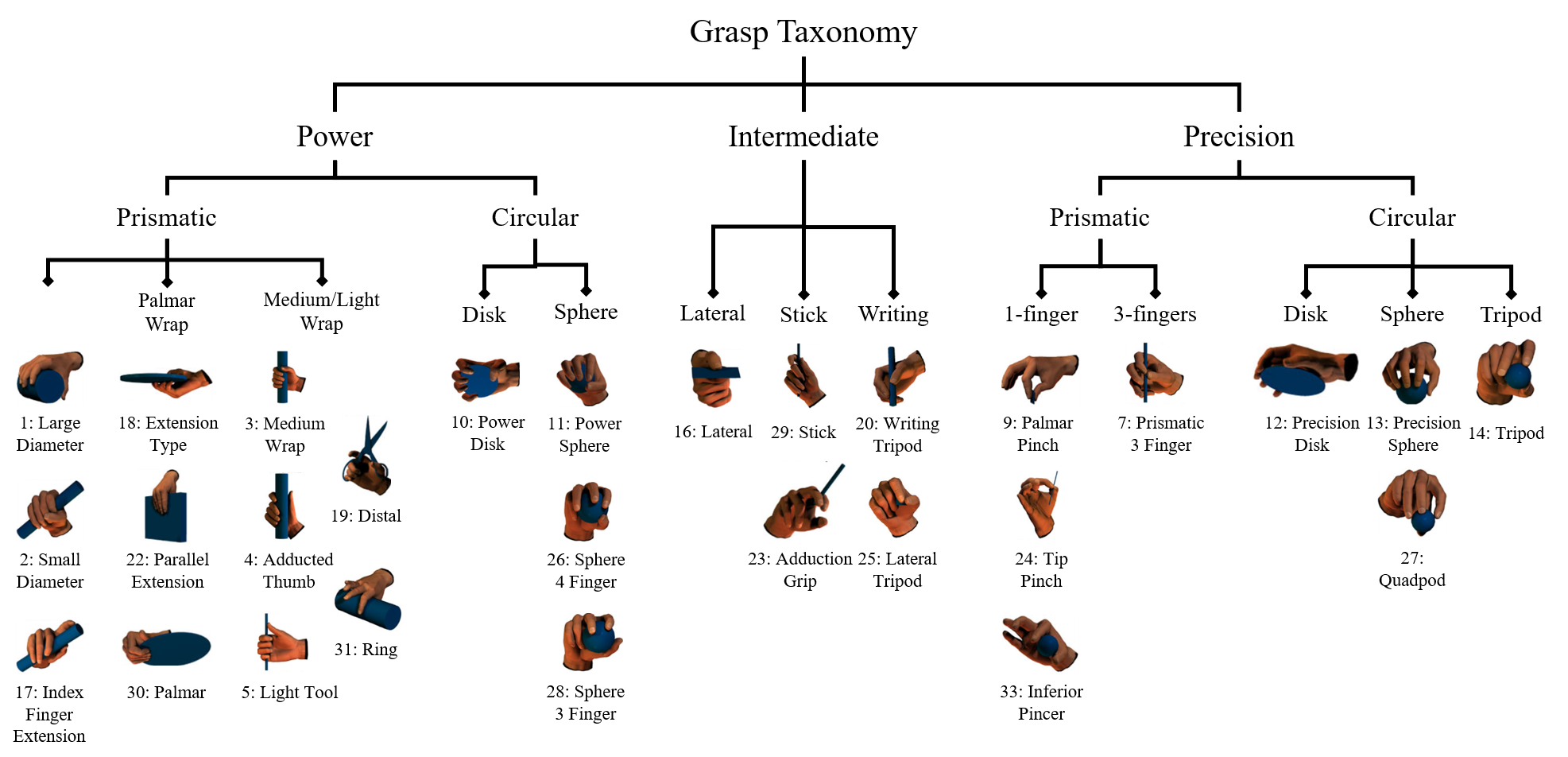}
  \includegraphics[height=0.28\linewidth]{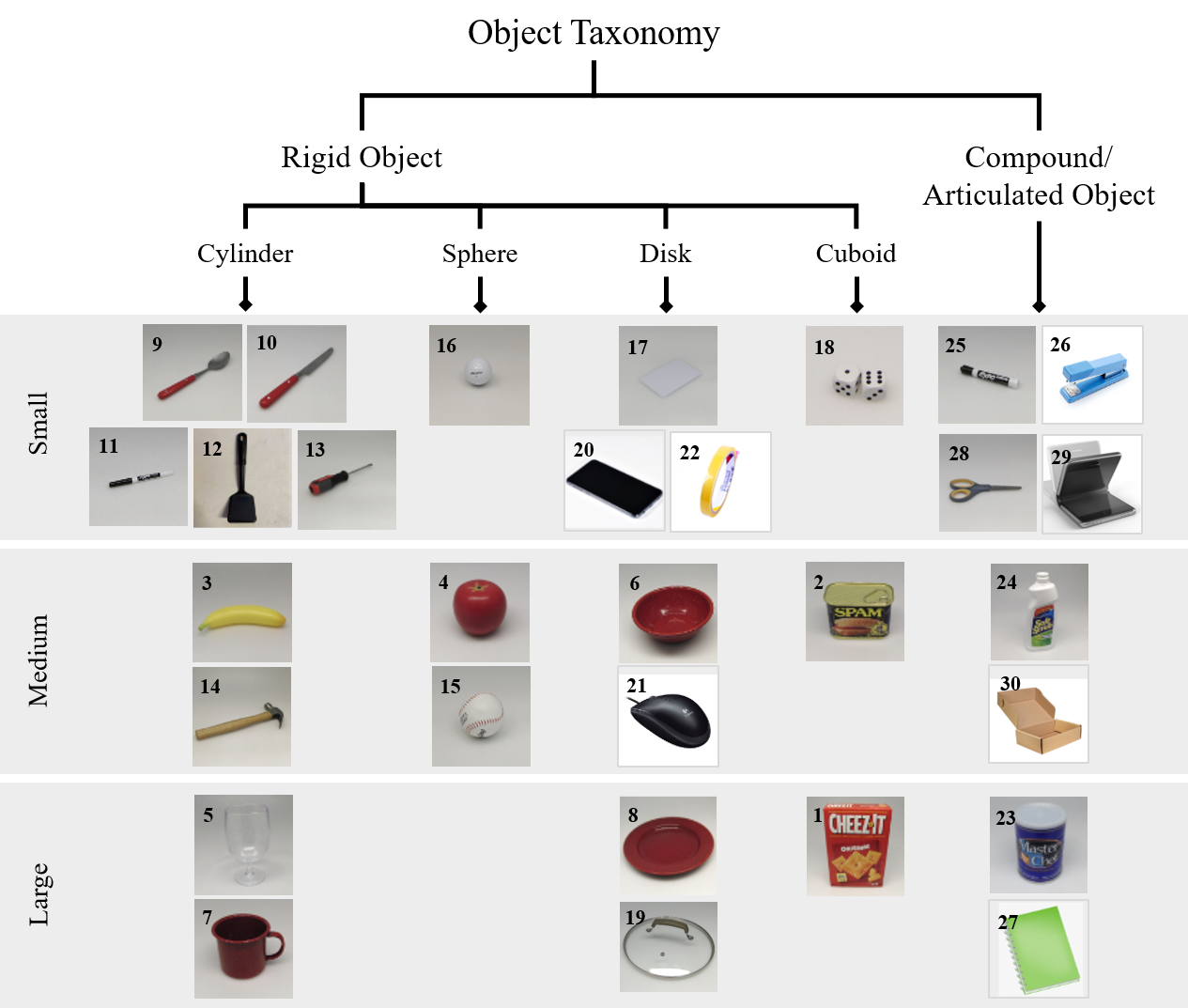}
  \caption{\textbf{(left) 33 hand grasping taxonomies, (right) 30 objects used in the dataset.} The object types are cylinder, sphere, disk, cuboid, or compound/articulated. They are further dividend to small/medium/large sizes, purporting to cover all grasp taxonomies. 
}
\label{fig:taxonomy}
\end{figure}

\begin{figure}[t]
  \centering
  \includegraphics[height=0.28\linewidth]{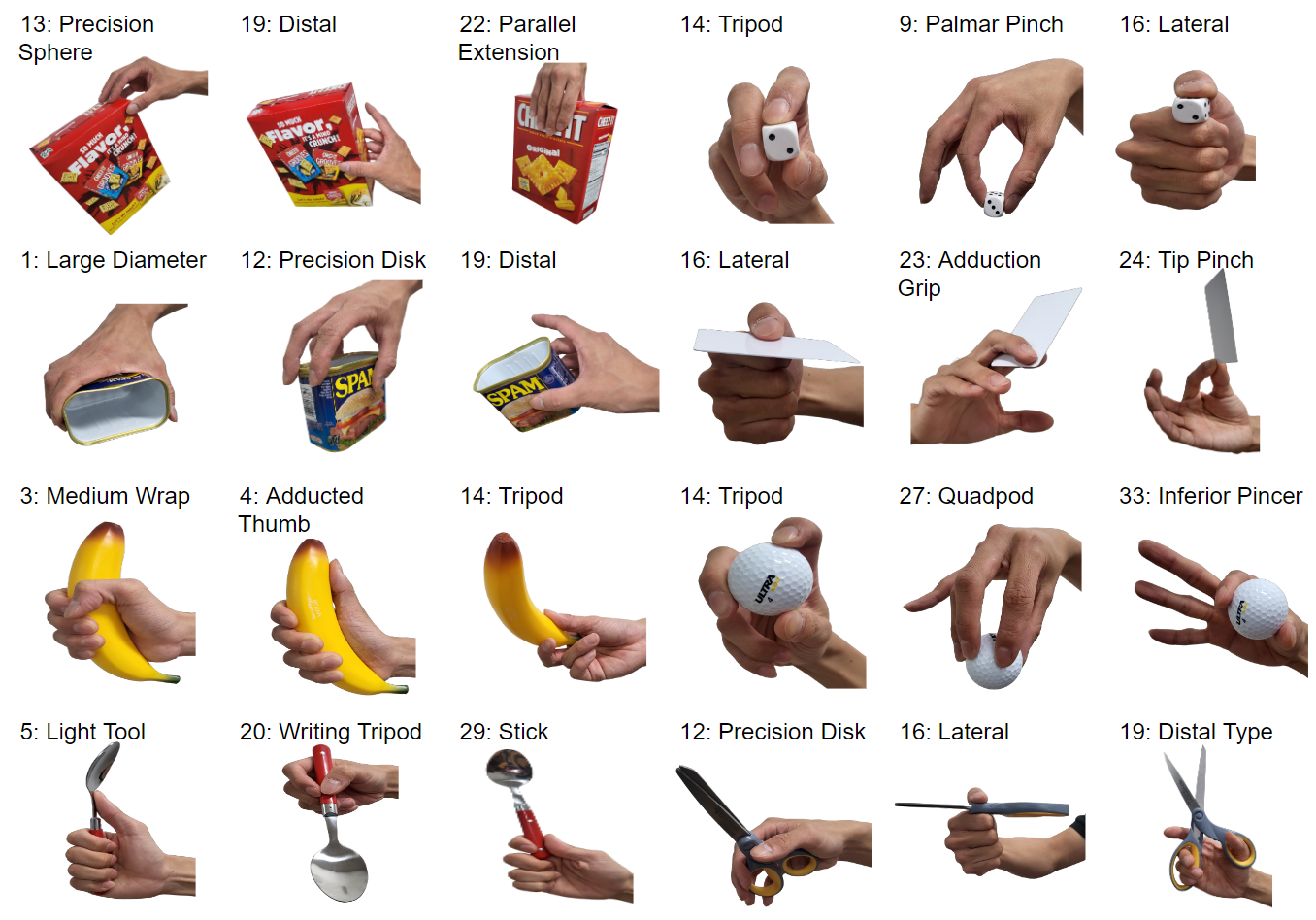}
  \includegraphics[height=0.28\linewidth]{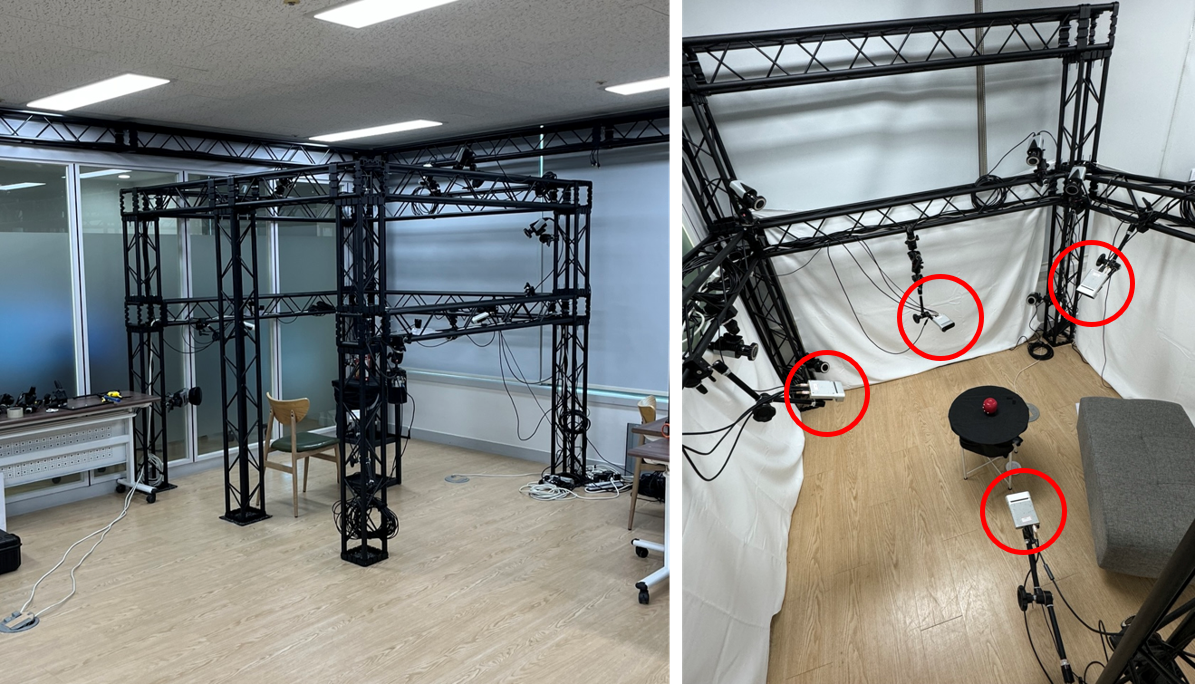}
   \caption{\textbf{(left) Per-object taxonomy examples (right) System setup.}
   The full list is shown in the supplementary.}
\label{fig:allobjects}
\vspace{-\baselineskip}
\end{figure}

While interacting with various objects, we often take specific grasp poses based on the object's shape and intention. To capture a wide range of hand pose space, especially to cover all grasping taxonomies, we identified 22 types of objects from the YCB dataset\cite{calli2015ycb} and 8 other daily objects. These objects are selected considering factors, such as the primitive shapes of objects (cylinder, sphere, disk, cube), especially grasp areas, object sizes, and additional articulated/compound objects (see Fig.~\ref{fig:taxonomy} (right)). Referring to previous studies on hand-object grasp taxonomy \cite{feix2015grasp, stival2019quantitative, liu2014taxonomy, arapi2021understanding, cini2019choice}, we captured the three most common grasp classes for each object. Example per-object taxonomies are shown in Fig.~\ref{fig:allobjects}. Also, some grasp classes out of 33 are seemingly redundant, visually hard to distinguish, and geometrically close; we redefined them to 28 grasp classes, with their indices following those presented in \cite{feix2015grasp}. Compared to the existing benchmarks (MOW\cite{cao2021reconstructing} and OakInk\cite{yang2022oakink}), we have a relatively smaller set of objects. However, we span more grasp space with large intra-class variations, thanks to diverse hand shapes and the number of frames. We consider synthetic object augmentation in textures and shapes beyond grasp areas and background augmentation as future work. 

\vspace{-0.4\baselineskip}

\vspace{-0.2\baselineskip}

\subsection{Hardware Setup and Data Collection}

\textbf{Sensors.} Fig.~\ref{fig:allobjects} shows the recording studio setup, where 4 temporally synchronized RGB-D cameras (Azure Kinect) are positioned around the designated space. The one in the backside is roughly at users' eye locations, imitating a fixed egocentric view. The cameras capture RGB and depth at 1920x1080 resolution and 30 FPS. For object poses, 8 IR cameras with a frame rate of 120 FPS were set, and 3 to 5 optical markers(3mm) were attached to each object. Notably, no markers were used on hands to maintain their realistic appearance, thereby minimizing the potential degradation of image features in networks trained using our dataset due to RGB image contamination (cf. ARCTIC\cite{fan2023arctic}). However, markers on objects can still limit hand poses, so we minimized this impact by placing markers on regions least likely to be grasped (e.g., the blade of scissors). Note that all objects were symmetrical enough to place markers while avoiding contact areas. Temporal synchronization between the RGB-D and IR cameras was obtained by manually aligning the starting frames during each recording session through a start blink of the LED.

\noindent \textbf{Data acquisition.} We conducted data capture involving 99 participants with diverse hand sizes, shapes, and textures. Detailed instructions regarding grasp classes for each object were provided, and participants were requested to grasp each object with their right hand according to the specified grasp while freely performing pose variations such as translation and rotation. Each participant completed the procedure 2 to 4 times, with each trial recorded for 20 seconds to adequately capture actions ranging from reaching for the object to freely manipulating it in the air and eventually placing it back down. This way, diverse intra-class variations were captured.

\begin{figure}[t]
  \centering
  \includegraphics[width=0.70\columnwidth]{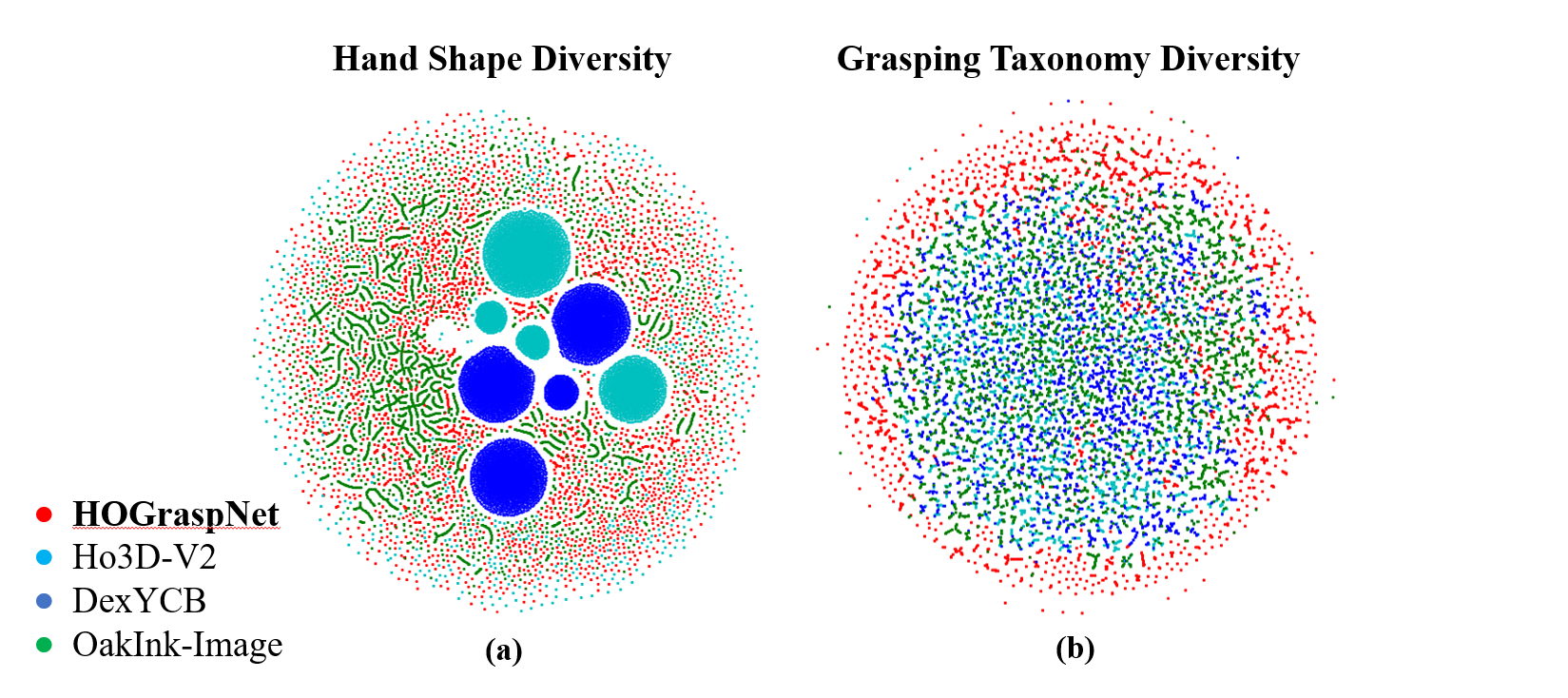}
  \vspace{-0.3\baselineskip}
   \caption{\textbf{t-SNE~\cite{van2008visualizing} visualization of (left) MANO~\cite{romero2022embodied} shape parameter distributions and (right) grasp feature distributions.}}
   \label{fig:tsne}
   \vspace{-\baselineskip}
\end{figure}

\vspace{-0.2\baselineskip}
\subsection{Data Distributions}
\label{subsec:data_distributions}

To further demonstrate that our dataset captures more comprehensive hand grasps, we visualize our data distribution in comparison to HO3D~\cite{hampali2020honnotate}, DexYCB~\cite{chao2021dexycb}, and OakInk~\cite{yang2022oakink}. In Fig.~\ref{fig:tsne}\textcolor{red} {(a)}, we show t-SNE~\cite{van2008visualizing} visualizations of the four datasets in the MANO~\cite{romero2022embodied} shape parameter space. Our dataset captures more hand shape diversity than the others, as a larger number of hand identities were included (as shown in Tab.~\ref{tab:comparisionHandObject}). Fig.~\ref{fig:tsne}\textcolor{red}{(b)} shows the t-SNE visualizations in the grasp feature space. For grasp feature extraction, we train a hand auto-encoder with mesh reconstruction loss and the auxiliary contact reconstruction and grasp classification losses (see Sec.~\ref{subsec:grasp_classification} for more details) to obtain features that capture hand pose and grasp configurations. Ours is shown to be significantly more diverse than the other datasets in this feature space as well, thanks to our data acquisition process associated with carefully determined grasping types. We hope that the comprehensiveness of our dataset can serve as an effective prior for the downstream tasks related to hand-object interaction.

\vspace{-0.2\baselineskip}
\subsection{MANO and Object Annotation}
\label{subsec:annotations}

For MANO~\cite{romero2022embodied} hand and object annotation, we use an automatic annotation and verification pipeline inspired by prior studies~\cite{hampali2020honnotate,chao2021dexycb}. In the following subsections, we discuss each step of our pipeline, while more details can be found in the supplementary.

\noindent \textbf{Data Preprocessing.}
We downsample the captured RGB-D frames from 30FPS to 10FPS to filter temporally redundant samples. We also prepare the hand and object segmentation masks using the DeepLabv3~\cite{chen2017rethinking} model, which is fine-tuned using our data with a few manually annotated segmentation masks for each object.

\noindent \textbf{Initial Hand Keypoint Estimation.}
To prepare the initial hand keypoints used for MANO fitting, we use MediaPipe~\cite{lugaresi2019mediapipe} hand pose estimator, which is known to have high generalization ability. We estimate 2.5D hand keypoints from each multi-view frame and lift them to 3D keypoints via triangulation. However, such keypoint estimates may be noisy for viewpoints with high hand-object occlusion. To overcome this, we introduce a novel bootstrapping procedure to achieve a better 3D keypoint lifting quality. Given the 2.5D keypoint estimates from our four viewpoints $\{\textit{vp}_i\}_{i=0, 1, 2, 3}$, we obtain the lifted 3D keypoints $\{\hat{J}_i\}_{i=0, 1, 2, 3}$, where $\hat{J}_i \in \mathbb{R}^{21 \times 3}$ denotes 3D keypoints lifted using three viewpoints while \emph{excluding} $\textit{vp}_i$. We assume that if the MPJPE between (1) $\hat{J}_i$ projected onto $\textit{vp}_i$ and (2) the original 2.5D keypoint estimates from $\textit{vp}_i$ is above a threshold $\tau$, then the original estimates from $\textit{vp}_i$ is an outlier. In this way, we filter out noisy 2.5D keypoints during 3D lifting procedure to obtain more robust 3D keypoints per frame. The valid hand poses for each viewpoint and the visibility $v_i$ for each joint $i$ (computed using depth maps) are also stored to serve as pseudo-ground truth (GT) data in the following steps. 

\begin{figure}[t]
  \centering
  \includegraphics[width=\columnwidth]{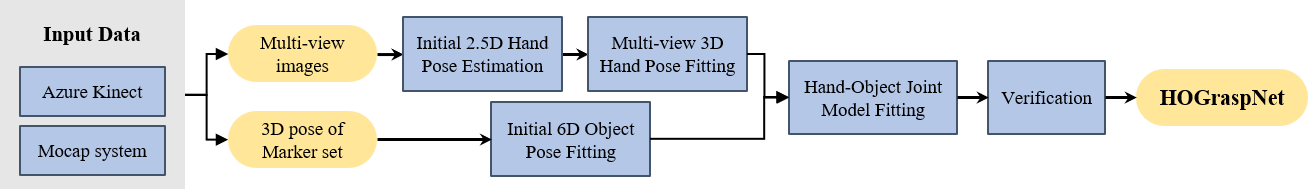}
   \caption{\textbf{MANO~\cite{romero2022embodied} and object annotation pipeline} (Section~\ref{subsec:annotations}).}
   \vspace{-\baselineskip}
\end{figure}

\noindent \textbf{Initial Object Pose Estimation.}
To obtain the initial 6D poses of an object, we attach optical sensors to the predefined surface locations of each object. Using multiple IR cameras, the 3D positions of each optical sensor are collected through in-built software. Then, the object's 3D rotation and translation are computed via Least-Squares Fitting to the marker positions.

\noindent \textbf{Multi-view Multi-frame Gradual Hand-object Model Fitting.}
In this stage, our goal is to fit MANO~\cite{romero2022embodied} hand and object template models to multi-view RGB-D frames and the initial hand and object poses. To this end, we formulate an optimization-based fitting scheme similar to previous works\cite{zimmermann2019freihand, chao2021dexycb}. To avoid local minima, we further propose to gradually fit the MANO parameters, such that our optimization consists of three stages: (1) fitting global hand transformation, (2) fitting partial hand poses extended from the wrist, and (3) fitting the full hand and object pose (see the supplementary for details).

Our overall loss function for the MANO pose $\theta \in \mathbb{R}^{48}$ and shape $\beta \in \mathbb{R}^{10}$ parameters and the object 6D pose $\phi \in \mathbb{R}^{6}$ can be written as:

\vspace{-1.2\baselineskip}

\begin{equation}\label{eq:loss}
\mathcal{L} = \lambda_{h}^{2D}\mathcal{L}_{h}^{2D} + \lambda_{o}^{3D}\mathcal{L}_{o}^{3D} + \lambda_{seg}\mathcal{L}_{seg} + \lambda_{depth}\mathcal{L}_{depth} +  \lambda_{reg}\mathcal{L}_{reg} + \lambda_{phy}\mathcal{L}_{phy}.
\end{equation}

\noindent $\mathcal{L}_{h}^{2D}$ measures the L2 distance between the pseudo GT 2D joints and the 2D projection of the MANO 3D joints weighted by the visibility $v_i$. $\mathcal{L}_{o}^{3D}$ computes the L2 distance between the 3D marker positions and the corresponding vertex position of an object model. $\mathcal{L}_{seg}$ and $\mathcal{L}_{depth}$ measures the L1 distance between the GT and the rendered segmentation masks and depth maps, respectively.

Following \cite{zimmermann2019freihand}, we also incorporate a regularization term $\mathcal{L}_{reg} = ||\tilde{\theta}||_{2} + ||\tilde{\beta}||_{2} + ||\theta_t - \theta_{t-1}||_{2} + ||\beta_t - \beta_{t-1}||_{2}$, which (1) penalizes MANO pose and shape parameters that deviate too much from the mean zero vectors and (2) encourages the previous and current hand parameters to be close for temporal consistency.
To additionally regularize the fitted hand and object meshes to be physically plausible, we incorporate another regularization term $\mathcal{L}_{phy}$, which is designed as a weighted sum of penetration loss and contact loss: $\mathcal{L}_{phy} = \lambda_{pen} \mathcal{L}_{pen} + \lambda_{contact} \mathcal{L}_{contact}$. For penetration loss $\lambda_{pen}$, we use a vertex normal projection-based technique used in \cite{hampali2020honnotate}. For contact loss $\mathcal{L}_{contact}$, we minimize the distances between hand and object vertices below a distance threshold $\tau$ to encourage physical contact. In Equation~\ref{eq:loss}, $\{\lambda_i\}_{i=h,\, o,\, seg,\, depth,\, reg,\, phy}$ is a set of scalar values to control the weighting between the loss terms. Please see the supplementary for more details about the annotation procedure.

\noindent \textbf{Post-verification.}
We conduct both automatic and manual verification steps to further filter out the noisy annotations. We compute the Intersection over Union (IoU) between the pseudo-GT and the rendered segmentation masks, filtering out annotations with an IoU below 0.6 in any view. Subsequently, we perform manual verification through crowdsourcing using LabelOn(\url{https://www.labelon.kr/}). Each crowdsourcer identifies misprocessed data that significantly deviates from the hand and object meshes or results from operational errors.

\vspace{-0.8\baselineskip}

\subsection{HALO Annotation}\label{sec:HALOsection}

\begin{figure}[t]
  \centering
\includegraphics[width=0.85\columnwidth]{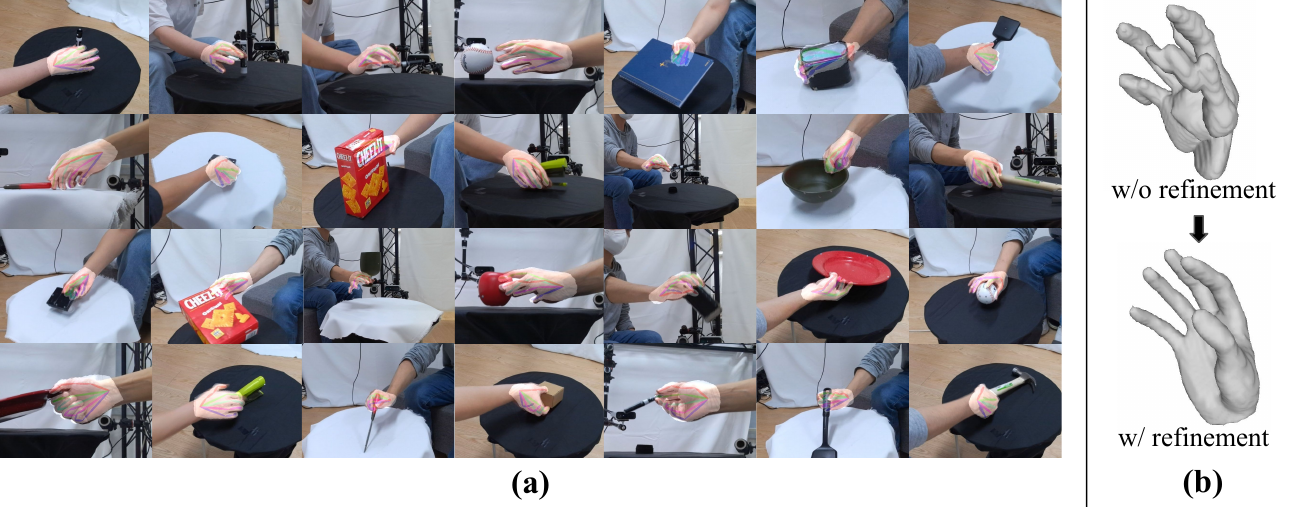}
  \vspace{-0.7\baselineskip}   \caption{\textbf{HALO~\cite{karunratanakul2021skeleton} fitting results.} \textbf{(a)} Annotated HALO hand examples. \textbf{(b)} Comparisons between the HALO shapes with and without applying inverse kinematics-based keypoint refinement~\cite{chen2023gsdf}.}
   \label{fig:halo_fitting_results}
\vspace{-0.6\baselineskip}
\end{figure}

\noindent \textbf{HALO~\cite{karunratanakul2021skeleton} fitting.} For hand shape annotation, we additionally provide the hand implicit surface based on HALO~\cite{karunratanakul2021skeleton}, which is a neural implicit representation that parameterizes an articulated occupancy field~\cite{mescheder2019occupancy} with 3D hand keypoints. Thus, a straightforward approach to fit HALO to our collected data would be to use the 3D hand keypoints lifted from the multi-view 2D keypoint estimates (as described in Section~\ref{subsec:annotations}) as an input to the HALO model. However, we observe that it leads to a less plausible implicit hand surface since the keypoints are not guaranteed to form a valid kinematic structure of the hand. Thus, we postprocess the lifted keypoints to the nearest keypoints on the hand space learned by MANO via the inverse kinematics algorithm in \cite{chen2023gsdf}. Our HALO fitting results are shown in Fig.~\ref{fig:halo_fitting_results}. 

\begin{figure}[t]
  \centering
  \includegraphics[width=0.85\columnwidth]{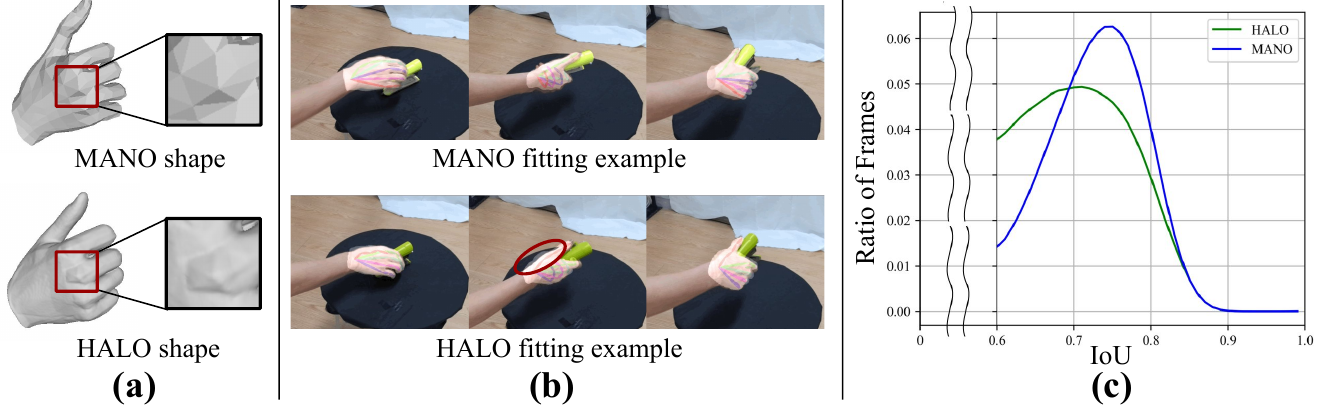}
  \vspace{-0.8\baselineskip}
   \caption{\textbf{Comparisons between HALO~\cite{karunratanakul2021skeleton} and MANO~\cite{romero2022embodied} fitting results.} \textbf{(a)} Hand shapes. \textbf{(b)} Fitting examples. \textbf{(c)} IoU distributions after post-verification stage.
   }
   \label{fig:halo_mano_comparisons}
\vspace{-0.8\baselineskip}
\end{figure}

\noindent \textbf{Comparisons with MANO~\cite{romero2022embodied} Fitting.}
As HALO~\cite{karunratanakul2021skeleton} is an occupancy function that takes hand keypoints as input, it does not require many hyperparameters (except for an occupancy threshold~\cite{mescheder2019occupancy}) for model fitting, while MANO fitting typically requires numerous loss weighting terms (i.e., $\lambda_{*}$ in Equation~\ref{eq:loss}) for defining the optimization objective. Thus, HALO can be more convenient and scalable for annotating a large-scale dataset. Also, HALO can produce hand shapes in a resolution higher than
MANO due to its resolution-independent nature (see Fig.~\ref{fig:halo_mano_comparisons}\textcolor{red}{(a)}). However, HALO is shown to capture less hand shape variation than MANO (see the red circle in Fig.~\ref{fig:halo_mano_comparisons}\textcolor{red}{(b)}) due to its keypoint-based parameterization for hand shape. As shown in Fig.~\ref{fig:halo_mano_comparisons}\textcolor{red}{(c)}), the IoU distribution of HALO is marginally worse than that of MANO, but it still achieves comparable mean IoU results (MANO: 0.739, HALO: 0.719) despite its simple annotation procedure.

\section{Experimental Results}
\label{sec:experiments}

In this section, we first report the split protocols of HOGraspNet (Section~\ref{subsec:split_protocols}). We then present our experimental results on grasp classification (Section~\ref{subsec:grasp_classification}) and hand-object pose estimation (Section~\ref{subsec:ho_pose_estimation}) using our dataset.

\subsection{Split Protocols}
\label{subsec:split_protocols}

For the evaluation setup, we generated five distinct train/test splits based on key components within our dataset:
\vspace{-0.4\baselineskip}
\begin{itemize}
\item{\textbf{S0 (default).} This split encompasses all subjects, views, objects, and grasp classes. The dataset is split by sequences, with the first sequence of each subject selected as the test set and the remaining sequences used for training.}
\item{\textbf{S1 (unseen subjects).} The dataset is split by subjects, following a 7:3 train/test ratio.}
\item{\textbf{S2 (unseen views).} The dataset is split by camera views, following a 3:1 train/test ratio.}
\item{\textbf{S3 (unseen objects).} The dataset is split by objects. 7 objects that collectively represent all 28 grasps are selected as the test set, while the other 23 objects are used for training.}
\item{\textbf{S4 (unseen taxonomy).} The dataset is divided by the grasping taxonomy. All the \textit{intermediate} grasp types in Fig.~\ref{fig:taxonomy} are selected as the test set, while others are used for training.}
\end{itemize}
\vspace{-0.4\baselineskip}
Note that we will release the exact split configurations through code. Also, refer to the supplementary for the benchmarking hand-object reconstruction results for each split.

\vspace{-0.4\baselineskip}
\subsection{Grasp Classification}
\label{subsec:grasp_classification}

\begin{figure}[t]
  \centering
  \includegraphics[width=0.95\columnwidth]{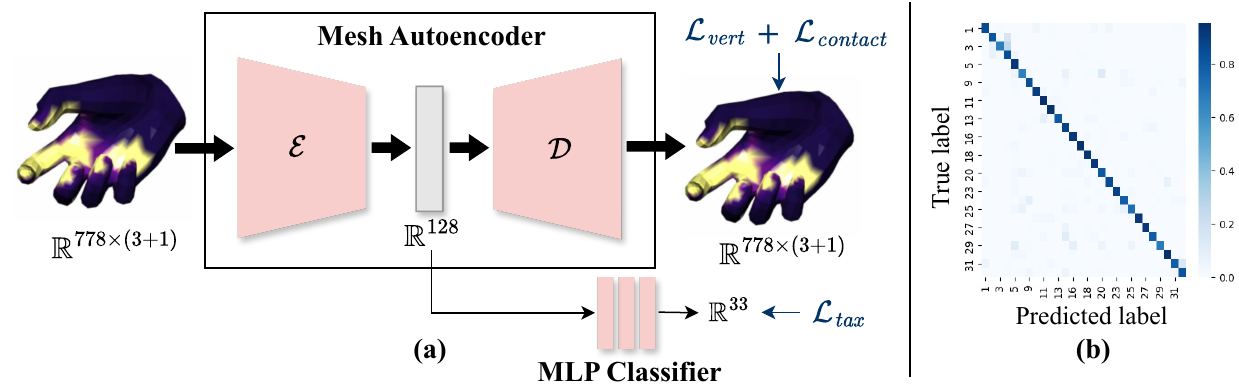}
  \vspace{-0.5\baselineskip}
   \caption{\textbf{(left) The network architecture for grasp classification, (right) confusion matrix.}}
   \vspace{-1.2\baselineskip}
   \label{fig:grasp_clas_model}
\end{figure}

We evaluate grasp classification performance on HOGraspNet using our S0 (Section~\ref{subsec:split_protocols}) benchmark setup. For the classification network, we modify the existing convolutional mesh autoencoder (CoMA~\cite{ranjan2018generating}) to take as input a hand mesh with per-vertex contact value as an additional vertex feature. The bottleneck feature of the autoencoder is fed to an MLP-based classifier to predict a grasp type. To train our model, we use L1 loss ($\mathcal{L}_{\mathit{vert}}$) that learns vertex reconstruction, and two cross-entropy losses that learn grasp taxonomy classification ($\mathcal{L}_{\mathit{tax}}$) and contact classification~\cite{grady2021contactopt} ($\mathcal{L}_{\mathit{contact}}$), where the range of contact value [0,1] is split into 10 bins. Our overall network architecture is shown in Fig.~\ref{fig:grasp_clas_model}. Note that we utilize auxiliary reconstruction losses for the classification task to obtain richer grasp features, which are also utilized for t-SNE visualization in Sec.~\ref{subsec:data_distributions}. Our model achieves 0.95 in f1 score for contact map reconstruction and 0.88 in accuracy for taxonomy classification. This experimental validation demonstrates that our grasping taxonomy can be delineated using hand meshes with contact maps without considering an object as input, indicating that our grasp annotation is generalized well across the samples.

\begin{figure}[t]
  \centering
  \includegraphics[width=0.95\columnwidth]{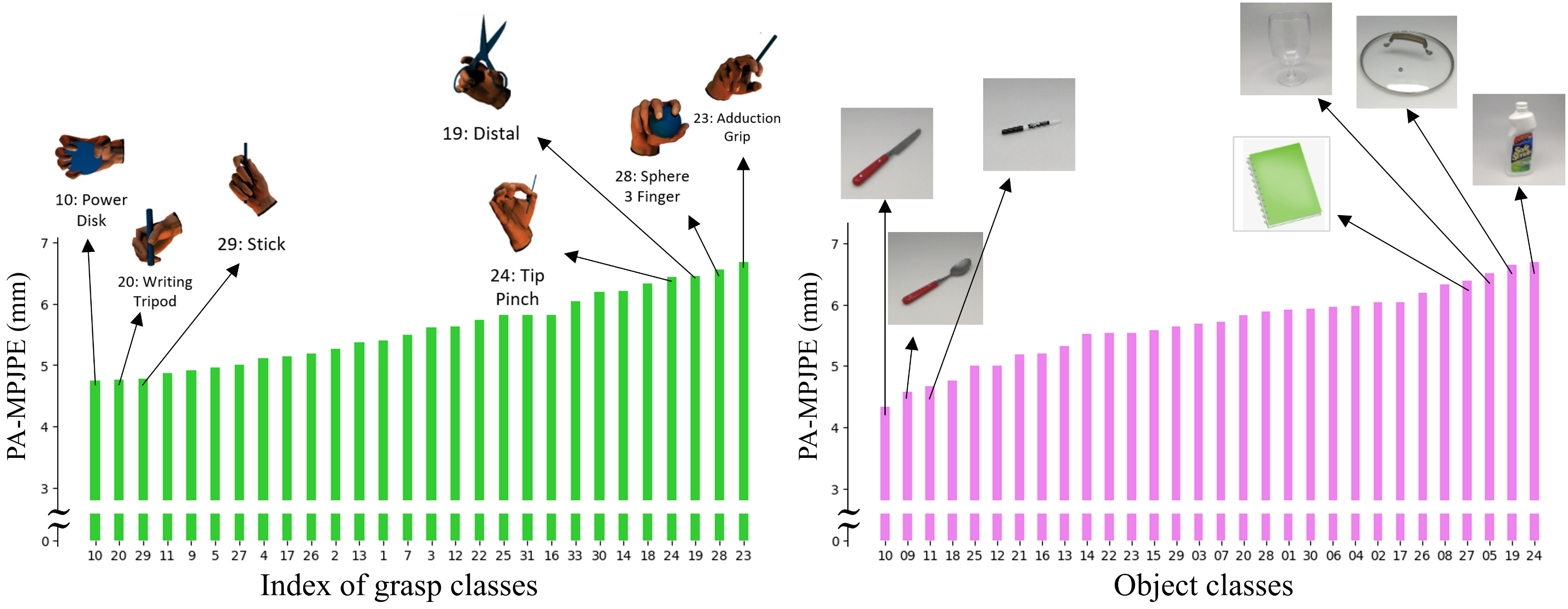}
  \vspace{-0.5\baselineskip}
   \caption{\textbf{Hand pose estimation results in PA-MPJPE (mm) (left) per grasp class and (right) object class.}}
   \label{fig:s0_split_nia}
\end{figure}

\vspace{-0.6\baselineskip}

\begin{table}[t]
\centering
\caption{\textbf{6D object pose estimation results in ADD-0.1D per object class using HOGraspNet.}}
\vspace{-0.2\baselineskip}
\resizebox{0.75\textwidth}{!}
{
\begin{tabular}{l|ccc}
\hline & 
\multicolumn{1}{c|}{ADD-0.1D} & \multicolumn{1}{c|}{}                      & ADD-0.1D
\\ \hline
1: cracker\_box   & \multicolumn{1}{c|}{88.79}      & \multicolumn{1}{c|}{16: golf\_ball}        & 46.27      \\
2: potted\_meat\_can  & \multicolumn{1}{c|}{59.47} & \multicolumn{1}{c|}{17: credit\_card}            & 34.06 \\
3: banana         & \multicolumn{1}{c|}{58.21}      & \multicolumn{1}{c|}{18: dice}              & 2.44       \\
4: apple          & \multicolumn{1}{c|}{75.74}      & \multicolumn{1}{c|}{19: disk\_lid}         & 99.29      \\
5: wine\_glass    & \multicolumn{1}{c|}{97.13}      & \multicolumn{1}{c|}{20: smartphone}        & 52.22      \\
6: bowl           & \multicolumn{1}{c|}{94.64}      & \multicolumn{1}{c|}{21: mouse}             & 41.40      \\
7: mug            & \multicolumn{1}{c|}{72.04}      & \multicolumn{1}{c|}{22: tape}              & 62.34      \\
8: plate          & \multicolumn{1}{c|}{99.42}      & \multicolumn{1}{c|}{23: master\_chef\_can} & 88.75      \\
9: spoon              & \multicolumn{1}{c|}{50.60} & \multicolumn{1}{c|}{24: scrub\_cleanser\_bottle} & 89.80 \\
10: knife         & \multicolumn{1}{c|}{38.17}      & \multicolumn{1}{c|}{25: large\_marker}     & 34.59      \\
11: small\_marker & \multicolumn{1}{c|}{28.29}      & \multicolumn{1}{c|}{26: stapler}           & 61.40      \\
12: spatula       & \multicolumn{1}{c|}{67.16}      & \multicolumn{1}{c|}{27: note}              & 88.70      \\
13: flat\_screwdriver & \multicolumn{1}{c|}{63.52} & \multicolumn{1}{c|}{28: scissors}                & 54.34 \\
14: hammer        & \multicolumn{1}{c|}{82.99}      & \multicolumn{1}{c|}{29: foldable\_phone}   & 25.02      \\
15: baseball          & \multicolumn{1}{c|}{73.72} & \multicolumn{1}{c|}{30: cardboard\_box}          & 77.80 \\ \hline
Avg               & \multicolumn{3}{c}{63.61}    \\ \hline
\end{tabular}
}
\label{tab:objectadd}
\vspace{-1.2\baselineskip}
\end{table}

\vspace{-0.4\baselineskip}

\subsection{Hand-Object Pose Estimation}
\label{subsec:ho_pose_estimation}

In this section, we present the benchmarking results on hand-object pose estimation on HOGraspNet using S0 split. We use HFL-Net~\cite{hfl2022} as a baseline, as it is the current state-of-the-art network on hand-object reconstruction. HFL-Net jointly estimates a hand mesh and 6D pose of an object from the input image via attention modules (please refer to \cite{hfl2022} and Sec.~\ref{sec:literature_survey} for more details).

In Fig.~\ref{fig:s0_split_nia}, we visualize the hand pose estimation results in PA-MPJPE for each grasp classes (left). The baseline achieves high accuracy across all grasp types with mean PA-MPJPE value of 5.67mm, which is comparable to the state-of-the-art hand pose estimation results on other widely-used hand-object datasets~\cite{fu2023deformer,xu2023h2onet,zheng2023hamuco,qu2023novel}. This further verifies that the quality of our hand annotation is decent, which allows for effective learning of the downstream hand pose-related task. We found that three of the top four grasp classes with the highest errors were not included in the DexYCB~\cite{chao2021dexycb} and HO3D~\cite{hampali2020honnotate} datasets, respectively. This implies that our dataset's newly introduced real grasp poses might be challenging for the hand pose estimation model trained on existing datasets. The supplementary materials provide details of the missing grasp classes per dataset. In Fig.~\ref{fig:s0_split_nia} (right), we also shows the hand pose metric per object classes. As expected, larger objects with more occlusion showed higher errors. Tab.~\ref{tab:objectadd} shows the object pose estimation results in ADD-0.1D per object class. We again achieve reasonable results that are comparable to the state-of-the-art object pose estimation performance on the other datasets~\cite{hfl2022, chao2021dexycb}, except for \emph{Dice} class. As the \emph{Dice} has small visible regions due to hand-object occlusions, increasing the ill-posedness of the pose estimation task. Additional results using other split protocols (S1-S4) can be found in the supplementary material.

\begin{table}[t]
\centering
\caption{\textbf{Cross-benchmark results on hand pose estimation using HFL-Net~\cite{hfl2022}.}}
\vspace{-0.8\baselineskip}
\resizebox{0.65\columnwidth}{!}{%
\begin{tabular}{cc|cc}
\hline
Train Set      & Test Set  & MPJPE (mm)       & PA-MPJPE (mm) \\ \hline
HO3D~\cite{hampali2020honnotate}        & DexYCB~\cite{chao2021dexycb}    & 57.31            & 10.31        \\
HOGraspNet & DexYCB~\cite{chao2021dexycb}    & \textbf{42.65}   & \textbf{9.36}\\
\hline
\end{tabular}
}
\label{tab:cross}
\end{table}

\begin{figure}[t]    
\centering
\includegraphics[width=\textwidth]{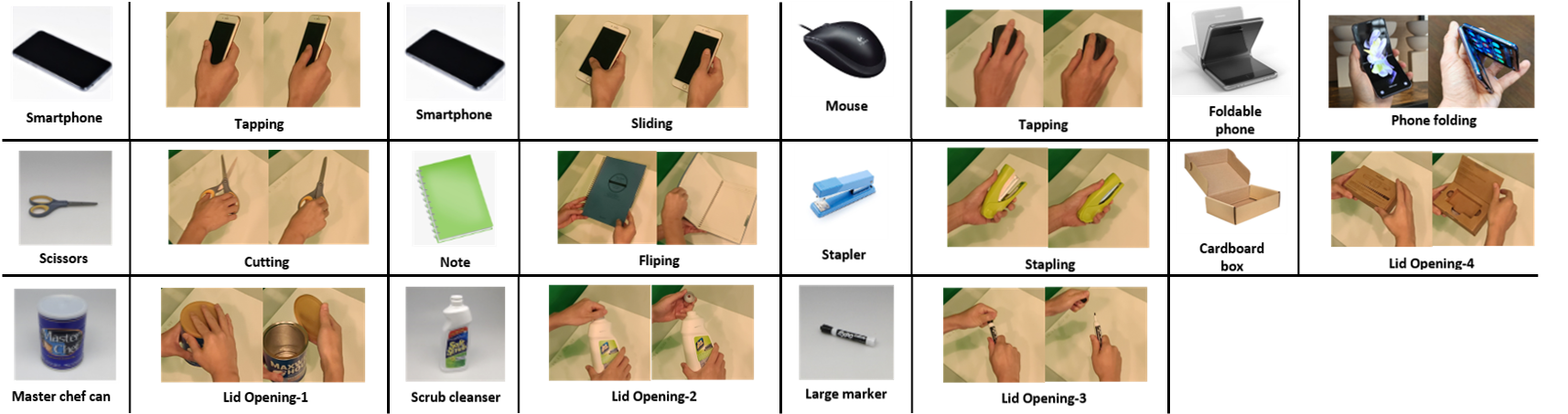}
\vspace{-0.8\baselineskip}
\caption{\textbf{Non-grasping action sequences in our dataset.}}
\label{fig:dynamic}
\end{figure}

\subsection{Cross-benchmark results on Hand Pose Estimation}

\noindent We additionally report the cross-validation results on hand pose estimation, following the experimental setup used in~\cite{yang2022oakink}. Since HFL-Net~\cite{hfl2022} is an object-aware network, we conducted experiments on samples with object classes that mutually exist in all the datasets to perform fair comparisons. In Tab.~\ref{tab:cross}, the network trained on HOGraspNet achieves better estimation accuracy than the network trained on HO3D~\cite{hampali2020honnotate}, indicating the comprehensiveness of HOGraspNet.

\section{Conclusions}

We have proposed a real RGB-D dataset, HOGraspNet, featuring comprehensive grasp labels. We have also presented the experimental results on grasp classification, and hand-object pose estimation. Our dataset captures diverse hand-object interactions involving 30 objects, 99 participants, and 90 interaction scenarios. It includes MANO~\cite{romero2022embodied} and HALO~\cite{karunratanakul2021skeleton} 3D hand meshes, 3D keypoints, object meshes, contact maps, and grasp annotations for every sequence. The benchmark notably improves accuracy across datasets by a broader range of interaction scenarios compared to the existing datasets.

\noindent \textbf{Limitations and Future Work.} We aimed to incorporate various compound and articulated objects to capture dynamic actions. However, we currently treat them as rigid objects. Nevertheless, interaction actions like tapping, folding, and opening have already been recorded, as illustrated in Fig.~\ref{fig:dynamic}. We plan to update the dataset with the object articulation annotations in the future. Furthermore, the dataset can be improved by including non-grasping actions such as pushing, throwing, squeezing, or deforming non-rigid objects like plastic bottles and sponges, which will be addressed as our future work.

\noindent \textbf{Acknowledgement} This work was in part sponsored by NST grant (CRC 21011, MSIT), IITP grant (No.2019-0-01270 and RS-2023-00228996, MSIT).

% ---- Bibliography ----
%
% BibTeX users should specify bibliography style 'splncs04'.
% References will then be sorted and formatted in the correct style.
%
\bibliographystyle{splncs04}
\bibliography{egbib}

\clearpage

\end{document}